# Le sens de la famille : analyse du vocabulaire de la parenté par les plongements de mots

Ludovic Tanguy, Cécile Fabre, Nabil Hathout, Lydia-Mai Ho-Dac

CLLE : CNRS & Université de Toulouse –
{ludovic.tanguy, cecile.fabre, nabil.hathout, lydia-mai.ho-dac}@univ-tlse2.fr

## Abstract

In this study, we propose a corpus analysis of an area of the French lexicon that is both dense and highly structured: the vocabulary of family relationships. Starting with a lexicon of 25 nouns designating the main relationships (son, cousin, mother, grandfather, sister-in-law etc.), we examine how these terms are positioned in relation to each other through distributional analyses based on the use of these terms in corpora. We show that distributional information can capture certain features that organize this vocabulary (descent, alliance, siblings, genre), in ways that vary according to the different corpora compared.

## Résumé

Dans cette étude nous proposons une analyse en corpus d'une zone du lexique français à la fois dense et très structurée : le vocabulaire des relations familiales. En partant d'un lexique de 25 noms désignant les principales relations (fils, cousine, mère, grand-père, belle-sœur etc.), nous avons examiné la façon dont ces termes se positionnent les uns par rapport aux autres au travers des analyses distributionnelles basées sur l'emploi de ces termes en corpus. Nous montrons que l'information distributionnelle permet de capter certaines caractéristiques qui organisent ce vocabulaire (ascendance, alliance, germanité, genre), selon des modalités qui diffèrent dans les différents corpus que nous analysons.

**Mots clés :** noms de parenté, relations familiales, plongements de mots

## 1. Introduction

Les noms de parenté, qui constituent des objets privilégiés de l'anthropologie, ont également intéressé la linguistique du fait de la grande richesse terminologique qu'ils manifestent à travers les langues, et de leurs propriétés syntaxiques et pragmatiques particulières (Takagaki 2010, Barque 2015). A ce jour, peu d'études de corpus se sont intéressées spécifiquement à ce vocabulaire d'un point de vue plus sémantique. On peut mentionner celles de Čermáková et Mahlberg (2021) qui s'intéressent aux noms de famille selon l'angle du genre dans des textes littéraires pour enfants et qui montrent comment l'étude comparée des noms *père* et *mère* est révélatrice d'usages linguistiques et de représentations sociales très contrastées. Aujourd'hui, les outils d'analyse distributionnelle permettent d'étudier à grande échelle l'organisation sémantique de classes de mots, à la manière de ce qu'ont montré Huyghe et Wauquier (2020) sur les noms d'agent. En proposant une représentation condensée basée sur les cooccurrences de chaque mot dans un corpus, ils permettent de définir simplement une mesure de similarité qui indique pour deux mots du texte leur propension à apparaître dans les mêmes contextes, et de ce fait leur proximité sémantique selon l'hypothèse distributionnelle. Notre objectif est d'appliquer ce type d'approche au vocabulaire de la parenté en calculant les représentations distributionnelles de ces mots (*embeddings* ou plongements de mots) pour en capter l'organisation sémantique et l'observer à travers différents corpus, susceptibles d'illustrer des modes de représentation contrastés de la famille. Nous cherchons en particulier à savoir si les catégorisations usuelles en termes de relations d'alliance, de descendance, de germanité (relations frère-sœur) ou la dimension du genre sont captées par l'information distributionnelle et s'avèrent structurantes dans les corpus. Nous produisons deux mesures de distance entre les termes de la famille : une distance distributionnelle calculée dans les différents modèles construits à partir de cinq corpus, et une distance calculée à partir de ces traits structurants. Nous montrons l'existence d'une corrélation entre ces deux mesures, et de différences entre les corpus qui suggèrent en particulier un poids différent du genre ou de l'ascendance pour expliquer les regroupements effectués sur une base distributionnelle. L'étude que nous présentons est réalisée sur le français, et s'inscrit dans une représentation traditionnelle de la famille occidentale.

## 2. Matériel : amorces, corpus et modèles

Notre démarche consiste tout d'abord à calculer, sur la base d'un échantillon de noms de parenté, la représentation vectorielle de chaque nom au sein de modèles distributionnels calculés à partir de plusieurs corpus.

### 2.1. Sélection d'un échantillon de noms de parenté

Nous avons sélectionné une liste de noms de parenté en partant d'un recensement disponible sur Wikipédia[1]. Cette liste initiale a été réduite à 25 noms (1) après élimination des termes dont la fréquence est inférieure à un seuil fixé à 100 pour chacun des corpus de l'étude. Ce seuil de fréquence permet d'obtenir des représentations correctes et stables pour chaque mot en termes de plongements lexicaux. Les termes écartés correspondent aux relations les plus éloignées (ex : *arrière-grand-mère*, *petite-nièce*, *demi-frère*, etc.) ou à des variantes moins usitées (*gendre* est bien plus fréquent que *beau-fils*, et *belle-fille* que *bru*). Nous avons par ailleurs choisi d'écarter le terme *femme* (en tant que correspondant féminin de *mari*) dont le sens majoritaire dans l'ensemble des corpus est celui de 'individu de sexe féminin'. Sa prise en compte aurait faussé nos résultats. Nous avons en revanche conservé *fille* dont l'acception 'femme jeune' n'est pas dominante au vu des modèles.

> (1) *beau-frère, beau-père, belle-fille, belle-mère, belle-soeur, cousin, cousine, épouse, époux, fille, fils, frère, gendre, grand-mère, grand-père, mari, mère, neveu, nièce, oncle, père, petit-fils, petite-fille, soeur, tante*

### 2.2. Corpus de l'étude

Nous faisons l'hypothèse que l'organisation de ce vocabulaire diffère selon la nature des discours qui font référence à la famille. Nous avons donc réalisé notre étude sur plusieurs corpus afin de mesurer l'impact de différents genres textuels sur la distribution des termes cibles et sur leur représentation dans les espaces sémantiques.

#### 2.2.1. Choix des corpus

Notre choix de cinq corpus a été guidé par la disponibilité de données exploitables de qualité et de taille suffisantes pour garantir une fréquence minimale des termes afin que leurs représentations vectorielles soient fiables.

- **Frantext** : sous-ensemble de la base textuelle Frantext (Pierrel 2003) constitué de 514 romans français du XXe siècle, pour un total de 30 millions de mots.
- **FrWaC** (Baroni et al. 2009) : corpus extrait du Web francophone moissonné en 2009 (2,3 millions de pages, 1,6 milliard de mots). FrWaC a été très utilisé dans de nombreuses expérimentations nécessitant de grands volumes de texte. Son contenu est notoirement mal contrôlé et la répartition des genres textuels qui le composent n'est pas connue.

---

[1] https://fr.wikipedia.org/wiki/Relation_de_parenté

- **Le Monde** : collection d'articles du quotidien *Le Monde* constituée de l'ensemble des archives de la période 1991-2000, soit 505 000 articles de presse pour un total de 200 millions de mots.
- **Revues.org** : corpus constitué d'articles scientifiques issus de différentes revues de sciences humaines et sociales en accès ouvert hébergées sur le portail revues.org (désormais journals.openedition.org) géré par OpenEdition. Il contient le texte intégral de 62 000 articles scientifiques moissonnés en 2013 pour un total de 200 millions de mots.
- **Wikipédia** : corpus construit à partir du code HTML de la version française de Wikipédia en date de décembre 2018 qui contient plus 2 millions d'articles encyclopédiques et 720 millions de mots.

Nous disposons ainsi de larges échantillons de discours variés : textes fictionnels, journalistiques, scientifiques, encyclopédiques, complétés par un vaste corpus hétérogène. Cette variété va nous permettre de mesurer dans quelle mesure les représentations des termes de la parenté sont sensibles aux caractéristiques textuelles.

*2.2.2. Prétraitements*

À partir du texte brut de chaque corpus, nous avons extrait une version lemmatisée et neutralisée en genre grammatical du texte. Nous avons utilisé l'analyseur Stanza (Qi et al. 2020) version 1.3 avec le modèle UD_French-GSD pour la segmentation, l'étiquetage morphosyntaxique et une première lemmatisation. La lemmatisation produite par Stanza n'étant pas toujours de qualité suffisante (contrairement à l'analyse morphosyntaxique), nous l'avons contournée en utilisant la méthode proposée par Tanguy et Hathout (2007) appliquée au lexique flexionnel Morphalou 3 (ATILF 2023). La lemmatisation est en effet une étape critique dans notre approche pour identifier les mots-cibles mais surtout leurs cooccurrents. Pour les classes ouvertes (noms, verbes, adjectifs et adverbes), nous avons utilisé dans Morphalou le lemme correspondant à la forme de surface (ou prédit à partir de cette forme) pour la catégorie et les traits flexionnels proposés par Stanza. Pour les classes fermées et en cas d'absence d'une forme de classe ouverte compatible dans le lexique nous avons gardé la lemmatisation proposée par Stanza.

Afin de limiter le biais liés au genre des mots-cibles, la lemmatisation doit supprimer toutes les marques du genre grammatical. Le modèle GSD de Stanza lemmatise déjà les déterminants et les pronoms par la forme masculine. Pour la même raison, nous avons remplacé tous les noms propres par une chaîne générique "NAM" pour réduire les biais qu'ils induisent dans le calcul distributionnel, notamment du fait des prénoms. Par exemple, la phrase en (2)[2] est lemmatisée comme en (3).

(2)   Austin, le neveu d'Edie emménage chez sa tante et entame rapidement une relation avec Julie, la fille de Susan.

(3)   NAM, le neveu de NAM emménager chez son tante et entamer rapidement un relation avec NAM, le fille de NAM.

---

[2] Extraite de la Wikipédia : https://fr.wikipedia.org/wiki/Desperate_Housewives

TANGUY Ludovic, FABRE Cécile, HATHOUT Nabil, HO-DAC Lydia-Mai

*2.3 Modèles*

Nous avons construit des modèles vectoriels sémantiques (ou plongements de mots) pour les cinq corpus en utilisant l'algorithme *Skip Gram with Negative Sampling* (SGNS) rendu populaire par Word2vec (Mikolov et al. 2013). Nous avons utilisé la version implémentée dans la bibliothèque Gensim (Rehurek & Sojka 2011) avec comme paramètres : 300 dimensions, fréquence minimale de 100 occurrences, fenêtre de 5 mots, taux d'exemples négatifs 5, sous-échantillonnage $10^{-3}$, 5 epochs.

L'algorithme SGNS (comme toutes les méthodes basées sur des réseaux de neurones) fait intervenir un ensemble de processus aléatoires, ce qui entraîne un non-déterminisme et de ce fait une certaine instabilité des modèles vectoriels. Afin de limiter l'impact de ces aléas, il est d'usage de générer plusieurs modèles avec les mêmes données et paramètres, et de les utiliser conjointement (Pierrejean and Tanguy 2018). Nous avons donc, pour chaque corpus, entraîné 5 modèles et construit une représentation vectorielle en concaténant les 5 vecteurs de chaque mot. Les espaces dans lesquels les calculs présentés dans la suite ont été réalisés sont donc de dimension 1500 pour chaque corpus. Nous utilisons de plus un modèle *global* à 7500 dimensions dont les représentations sont la concaténation de celles des 25 modèles correspondant aux 5 corpus.

## 3. Les familles dans les modèles vectoriels

*3.1 Des familles très soudées*

Notre première approche a consisté à calculer, pour chacune des amorces sélectionnées, la liste des voisins distributionnels les plus proches. La proximité est mesurée classiquement en utilisant la similarité cosinus entre les vecteurs correspondants aux mots du vocabulaire. La première constatation est que, comme attendu, le lexique de la famille forme un ensemble cohésif et dense.

Dans les modèles des 5 corpus, et pour les 25 amorces, si l'on considère les 10 voisins les plus proches, 81% sont des mots de cette même liste d'amorces. Dans le modèle *global*, la cohésion des termes de parenté est plus forte encore et atteint 92,8%. Le score moyen pour les termes varie entre 95% pour *oncle* et 41,7% pour *fille*. Si l'on effectue une analyse par corpus cette fois, en faisant la moyenne sur les 25 amorces, le score le plus élevé est de 88% pour Revues.org et le plus bas 73% pour Frantext. Notons que la quasi-totalité des autres mots qui figurent parmi les 10 plus proches voisins sont liés à la famille. On y trouve certains termes écartés de la liste initiale (*petite-nièce*), des termes génériques sous-spécifiés (*parent*, *aïeul*), des qualificatifs (*aîné*, *veuve*, *adoptif*), des formes relevant d'un registre familier (*maman*, *mamie*) ou des fautes d'orthographes (*mere*, *frére*). Si l'on intègre ces termes, le taux moyen de mots relatifs à la famille parmi les 10 voisins les plus proches monte à 93%.

Bien entendu, le score diminue rapidement lorsque l'on refait le calcul à un rang plus élevé : 66,7% des voisins figurent en moyenne parmi les amorces au rang 25. Le mécanisme distributionnel fait en effet apparaître des termes relevant d'autres champs sémantiques qui restent à explorer.

*3.2 Organisation des familles*

Cette cohésion distributionnelle des termes de la famille se double d'une organisation interne à la classe. Afin d'observer leurs positions relatives, nous avons effectué une projection des vecteurs depuis les espaces des plongements vers un espace à deux dimensions. Nous avons retenu la méthode t-SNE (*T-distributed Stochastic Neighbor Embedding*), une méthode probabiliste couramment utilisée pour projeter les plongements de mots (Van der Maaten and Hinton 2008). Nous avons utilisé l'implémentation proposée dans la bibliothèque *Rtsne* de R sur la matrice de distance cosinus (1-*cos*) avec une perplexité de 5 (les autres paramètres ayant les valeurs par défaut).

A titre exploratoire, les figures 1 à 3 semblent indiquer que l'organisation distributionnelle du vocabulaire de la famille recouvre partiellement les catégories descriptives usuelles des liens de parenté, tout en variant notablement d'un corpus à l'autre. La figure 1, qui représente l'espace fusionnant tous les corpus, montre à la fois une proximité systématique entre les paires de noms (reliées sur le graphique) qui ne se distinguent que par le trait de genre (ex : *frère - soeur*), et des regroupements qui associent plus largement entre eux les termes relatifs à l'alliance (*belle-mère*, *gendre*, *belle-soeur*, etc.), l'ascendance (*mère*, *grand-père*, *tante*, etc.), la descendance (*fils*, *petite-fille*, *petit-fils*).

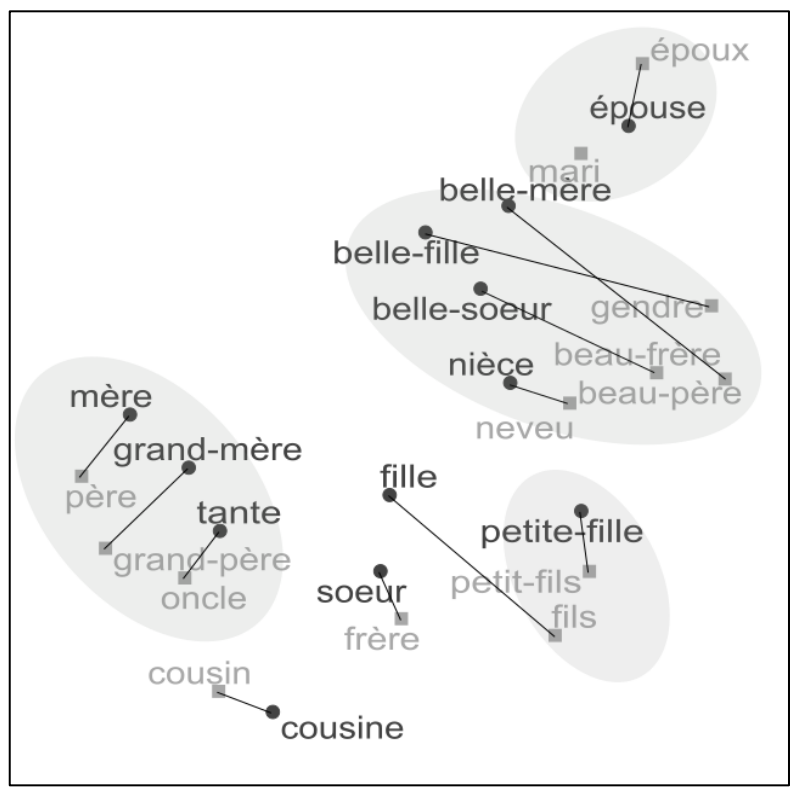

Figure 1 : Projection des distances distributionnelles par t-SNE :
Modèle global, marquage des mots par genre (gris foncé : féminin, gris clair : masculin), par paires genrées et par clusters identifiés manuellement

La figure 2 (à gauche) montre que pour Wikipédia le genre instaure une démarcation nette au sein de ce vocabulaire, alors que pour Frantext (à droite) la démarcation semble liée à l'ascendance (*beau-père* faisant exception, situé du côté des noms d'alliance). Dans les deux figures nous avons indiqué visuellement les traits correspondants pour chaque mot-cible.

TANGUY Ludovic, FABRE Cécile, HATHOUT Nabil, HO-DAC Lydia-Mai

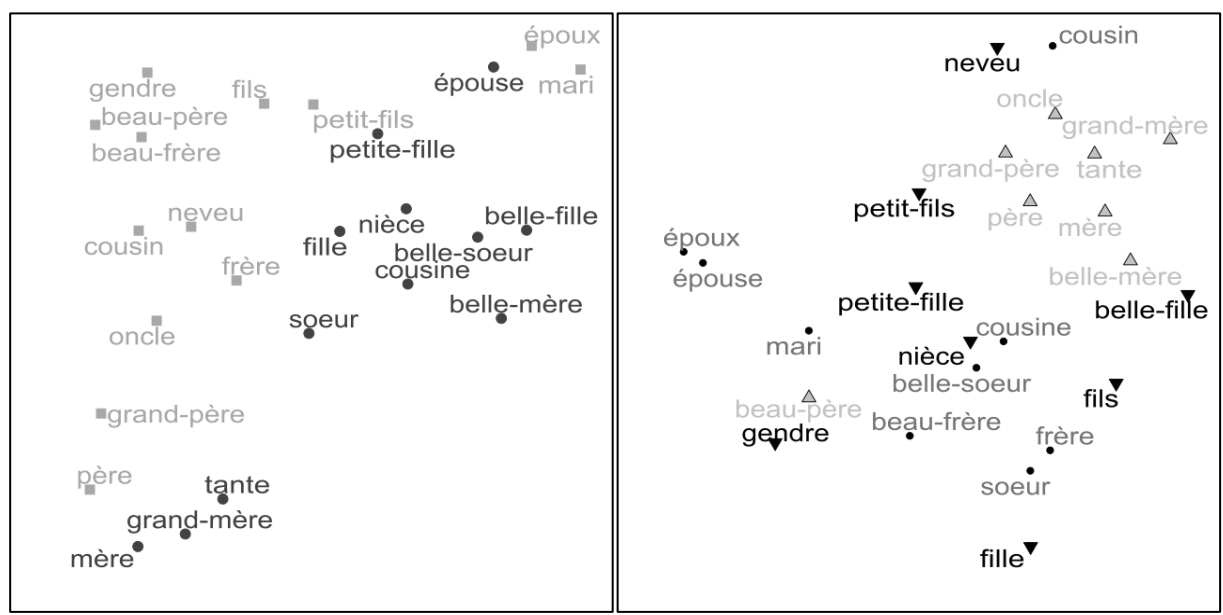

*Figure 2: Projection des distances distributionnelles par t-SNE.*
*A gauche : corpus Wikipédia, marquage par genre (gris foncé : féminin, gris clair : masculin).*
*A droite, corpus Frantext, marquage par génération (▲ : ascendants, ▼ : descendants)*

D'autres tendances apparaissent sur l'ensemble des figures (y compris celles non montrées ici). La forte polysémie anticipée de *fille* tend à l'isoler du groupe ou des différents clusters, et à l'éloigner de *fils*. Au contraire, certaines paires restent totalement inséparables quel que soit le corpus, notamment *époux*/*épouse*. Enfin, certaines relations relativement indirectes comme *cousin*, *cousine*, *nièce* et *neveu* auront tendance à se retrouver dans les zones médianes, ou à l'intersection de plusieurs groupements.

Il apparaît donc que les plongements de mots semblent suivre d'une certaine façon les propriétés pertinentes pour l'organisation des familles, mais avec des différences entre les corpus et donc a priori les genres textuels. Dans la section suivante nous vérifions et affinons cette hypothèse.

## 4. Adéquation entre décomposition et similarité distributionnelle

### *4.1. Décomposition en traits*

La description et la notation des relations entre les membres a fait l'objet de nombreuses propositions (Rivers 1910, Barry 2018 et les références qui y sont citées). Elle intéresse en premier lieu les anthropologues souhaitant décrire la diversité de ces relations dans les différentes sociétés humaines. Dans tous les modèles, les relations sont décrites relativement à Ego. Il existe essentiellement deux approches à la notation de ces relations. La notation traditionnelle décrit la relation entre deux membres de la famille par une séquence de relations élémentaires notées chacune par un symbole. Par exemple, le grand-père maternel d'Ego est noté MF dans la notation anglaise parce que c'est le père (F) de la mère (M) (*mother's father*). L'autre approche est l'analyse componentielle (Goodenough 1956, Augustins 2000) qui décrit la relation relativement à un ensemble de propriétés comme le genre, l'alliance (mariage, union), la filiation (qui peut être ascendante ou descendante), la génération, la germanité (appartenance à la même fratrie), la collatéralité (cousin, oncle), la consanguinité, etc. Les valeurs de ces traits peuvent être numériques comme pour la génération (+1 pour la génération des parents) ou booléennes comme pour la collatéralité (Vrai pour cousin) ou faire partie d'un ensemble comme pour le genre ({Féminin, Masculin, ...}).

Nous avons opté pour l'analyse componentielle qui permet de traduire plus simplement la distance entre les membres de la famille. Nous avons ainsi annoté les 25 termes d'amorce en utilisant les traits suivants : genre, ascendance, descendance, germanité, alliance. Le genre peut prendre deux valeurs : F ou M. Pour les autres traits, la valeur correspond au nombre de liens qui existent dans le chemin qui relie Ego à Alter (en pratique, ce nombre est soit 0 soit 1) : l'ascendance indique le nombre de liens ascendants (1 pour la mère ou le père) ; la descendance, le nombre de liens descendants (1 pour le fils ou la fille) ; germanité, l'existence ou non d'une relation entre germains (1 pour frère ou sœur) ; alliance, l'existence ou non d'une relation d'alliance (1 pour l'époux ou le beau-frère). Notons que certains noms de parenté sont ambigus comme par exemple *oncle* qui peut être frère de la mère, frère du père, mari de la sœur de la mère ou mari de la sœur du père. Dans ce cas, la valeur annotée est la moyenne des valeurs pour l'ensemble de ces chemins. Ainsi, la valeur d'alliance pour l'oncle est 0,5. Le tableau 1 donne le résultat de l'analyse componentielle pour quelques-unes des amorces.

| Amorce | genre | ascendance | descendance | germanité | alliance |
|---|---|---|---|---|---|
| *cousin* | M | 1 | 1 | 1 | 0 |
| *grand-mère* | F | 2 | 0 | 0 | 0 |
| *belle soeur* | F | 0 | 0 | 1 | 1 |
| *époux* | M | 0 | 0 | 0 | 1 |
| *oncle* | M | 1 | 0 | 1 | 0,5 |

*Tableau 1 : Caractérisation des amorces selon le modèle componentiel*

### 4.2. Adéquation entre la similarité distributionnelle et la décomposition en traits

Nous avons cherché à confronter la similarité distributionnelle entre deux mots de la famille à la distance entre ces deux mots selon l'analyse componentielle. Pour ce faire, nous avons effectué une régression linéaire multiple avec comme variable dépendante la similarité cosinus entre les vecteurs correspondant à chacune des amorces et comme variables indépendantes les différences mesurées sur les traits décrits précédemment. Plus précisément, nous avons mesuré pour chaque trait la valeur absolue de la différence entre les valeurs pour chaque mot de la paire. Pour les traits binaires (genre, germanité) la différence est de 0 si les traits sont identiques et de 1 s'ils sont différents.

Cela donne par exemple pour quelques exemples de paires (parmi les N=300 combinaisons de nos 25 amorces) les différences de traits indiquées dans le tableau 2.

TANGUY Ludovic, FABRE Cécile, HATHOUT Nabil, HO-DAC Lydia-Mai

| Paire d'amorces | genre | ascendance | descendance | germanité | alliance |
|---|---|---|---|---|---|
| *père - grand-mère* | 1 | 1 | 0 | 0 | 0 |
| *belle soeur - époux* | 1 | 0 | 0 | 1 | 0 |
| *fille - oncle* | 1 | 1 | 1 | 1 | 0,5 |

*Tableau 2 : Caractérisation des paires d'amorces en termes de traits différentiels*

Nous avons construit 6 modèles linéaires : un pour chacun des cinq corpus et un pour le modèle *global*. Dans chaque modèle, chaque différence de traits est négativement corrélée à la similarité distributionnelle (indiquant que toute différence de trait entre les mots de la paire correspond à une baisse de leur similarité) et est significative au seuil de 0,05. Le tableau 3 présente pour chaque corpus le $R^2$ ajusté (c'est-à-dire la part de variance de la similarité distributionnelle expliquée par les différences de traits) ainsi que, pour chaque trait, son importance relative mesurée en termes de perte relative de $R^2$ ajusté quand on l'élimine du modèle. Pour chaque modèle l'ensemble des variables atteint le seuil de significativité (p<0.05).

| Modèle | R² ajusté | genre | ascendance | descendance | germanité | alliance |
|---|---|---|---|---|---|---|
| Frantext | 0,238 | -0,29 | **-0,35** | -0,19 | -0,17 | -0,20 |
| FrWaC | 0,109 | -0,26 | -0,17 | -0,27 | -0,09 | **-0,40** |
| Le Monde | 0,375 | -0,18 | -0,25 | -0,24 | -0,09 | **-0,41** |
| Revues.org | 0,297 | **-0,33** | -0,16 | -0,25 | -0,15 | -0,30 |
| Wikipedia | 0,363 | **-0,50** | -0,34 | -0,18 | -0,06 | -0,10 |
| global | 0,350 | **-0,32** | -0,26 | -0,22 | -0,11 | -0,28 |

*Tableau 3 : Modèles linéaires par corpus - R² et perte relative par ablation des variables (valeur maximale par modèle en gras)*

Par exemple, dans le modèle linéaire construit sur le corpus Frantext, les différences de traits entre une paire de mots permettent d'expliquer 23,8% (R² ajusté = 0,238) de la variance observée sur les 300 paires dans la similarité distributionnelle entre les deux mots. Toujours dans Frantext, parmi les 5 variables prédictives (les 5 traits), c'est l'ascendance qui a le plus d'importance puisque retirer cette variable fait baisser le R² du modèle de 35% (validant l'interprétation de la figure 3 commentée en 3.2.).

On observe avant tout une importante variation du R² entre les corpus. C'est dans le corpus Le Monde (suivi de près par Wikipédia) que la similarité distributionnelle semble suivre au plus près les traits des mots de la paire. A l'inverse, dans FrWac cette similarité distributionnelle ne semble que très peu correspondre, 90% de la variance échappant au modèle. Parmi les différentes dimensions considérées, ce sont les différences de genre qui sont les plus importantes (et particulièrement pour Wikipédia, corroborant notre interprétation de la figure 2), alors que ce sont celles liées à l'implication d'une alliance qui ont le plus d'influence pour FrWac et Le Monde. Sur l'ensemble des modèles, les relations de descendance et de germanité sont les moins liées.

## 5. Conclusion

Nous avons présenté dans cet article une étude des termes de parenté et des relations familiales fondées sur les modèles vectoriels distributionnels. L'étude porte sur 25 noms que nous avons annotés relativement à cinq traits : genre, ascendance, descendance, germanité et alliance. Nous avons observé les représentations de ces noms dans des plongements construits à partir de 5 corpus. Nous avons d'une part constaté la très forte cohésion des termes de parenté : les voisins de ces termes sont eux-mêmes des termes de parenté. Nous avons d'autre part vu que les plongements de ces termes captent en partie les traits qui les caractérisent. Dans un second temps, nous avons (1) estimé dans quelle mesure les distances entre les termes dans les différents modèles rendent compte des différences entre leurs annotations et (2) déterminé l'importance relative de ces propriétés. Les résultats que nous avons obtenus sont conformes à nos constatations initiales. Par ailleurs, nous avons pu voir qu'il existait des disparités nettes entre les corpus qui reflètent des différences dans la façon dont ils rendent compte des positions relatives des termes dans l'espace sémantique. Si le genre est globalement le trait qui explique le mieux cette organisation, nous avons pu observer que c'est dans Wikipédia que la démarcation selon ce critère est la plus nette, alors que dans un corpus plus littéraire comme Frantext, l'ascendance est plus importante tandis que la similarité distributionnelle correspond davantage à l'alliance dans un corpus journalistique comme Le Monde.

Cette première exploration des termes de parenté a essentiellement porté sur les 25 noms et leur annotation. Les observations que nous en tirons à une échelle très globale nous encouragent à mener des analyses plus ciblées en corpus pour comprendre plus qualitativement ces différences de fonctionnement entre les discours. Par ailleurs, les modèles vectoriels nous offrent d'autres moyens de décrire l'organisation sémantique de ce vocabulaire, notamment par l'intermédiaire de la caractérisation des mots qui apparaissent dans les voisinages plus étendus des noms de parenté. En nous inspirant d'études comme celle de Huyghe et Wauquier (2020), nous envisageons également d'enrichir l'étude en considérant les traits sémantiques de ces voisins susceptibles de fournir d'autres informations pour expliquer l'organisation de ce vocabulaire (notamment les connotations). En ce qui concerne la comparaison entre corpus, il est possible d'appliquer des techniques de mesures des variations comme celles utilisées pour la détection de glissements sémantiques diachroniques (Tahmasebi et al. 2021). Nous souhaitons enfin profiter du caractère universel des relations de famille et réaliser le même type d'étude sur d'autres langues et/ou d'autres genres textuels.